\documentclass[onecolumn]{robbyant}           % Single column with abstract logo

\metadata[Website]{\url{https://technology.robbyant.com/lingbot-world-v2}}
\metadata[Github]{\url{https://github.com/robbyant/lingbot-world-v2}}
\metadata[Checkpoints]{\url{https://huggingface.co/robbyant/lingbot-world-v2}}
\newcommand{\methodname}{LingBot-World}
\newcommand{\methodold}{\texttt{\methodname}\xspace}
\newcommand{\methodaka}{\texttt{\methodname~2.0}\xspace}
\newcommand{\method}{\texttt{\methodname-Infinity}\xspace}

\newcommand{\E}{\mathbb{E}}

\newcommand{\Loss}{\mathcal{L}}

\title{Infinite Worlds with Versatile Interactions}

\author{
\begin{center}
    Zelin Gao,
    Qiuyu Wang,
    Jiapeng Zhu,
    Jingye Chen,
    Zichen Liu,
    Qingyan Bai,
    Jiahao Wang,
    \\[2pt]
    Yufeng Yuan,
    Hanlin Wang,
    Yichong Lu,
    Ka Leong Cheng,
    Haojie Zhang,
    Jian Gao,
    \\[2pt]
    Tianrui Feng,
    Yuzheng Liu,
    Yao Yao,
    Yinghao Xu,
    Xing Zhu,
    Yujun Shen,
    Hao Ouyang$^{\dagger}$
    \\[9pt]
    $^{\dagger}$Project Lead
\end{center}
}

\begin{document}
\abstract{
We present \methodaka (also known as \method), an advanced iteration of \methodold featuring four distinct upgrades.
(1) Our model achieves an \textbf{\textit{unbounded interaction horizon}} while maintaining consistent output quality, benefiting from a carefully crafted causal pretraining paradigm.
(2) Through distilling a real-time variant from the base model, our system guarantees \textbf{\textit{rapid response time}}, sufficient to drive 720p video streams at 60 fps.
(3) Compared to the previous version, this update introduces \textbf{\textit{highly diverse interactive elements}}, comprising a broader spectrum of actions (\textit{e.g.}, attacking, archery, spell-casting, and shooting) alongside a richer variety of text-driven events.
(4) We pioneer the integration of an \textbf{\textit{agentic harness}} within the domain of world modeling, wherein a pilot agent is tasked with planning and executing character behaviors, while a director agent is responsible for synthesizing novel environmental elements as the scene progresses.
Additionally, to facilitate a shared experience, we develop an interface that permits multiple players to simultaneously immerse themselves in this vivid world simulator.
We pair our primary 14B model with a lightweight 1.3B counterpart, which supports effortless deployment on a single GPU.
}

\maketitle
\begin{figure}[!h]
\centering
\vspace{-12pt}
\includegraphics[width=0.98\linewidth]{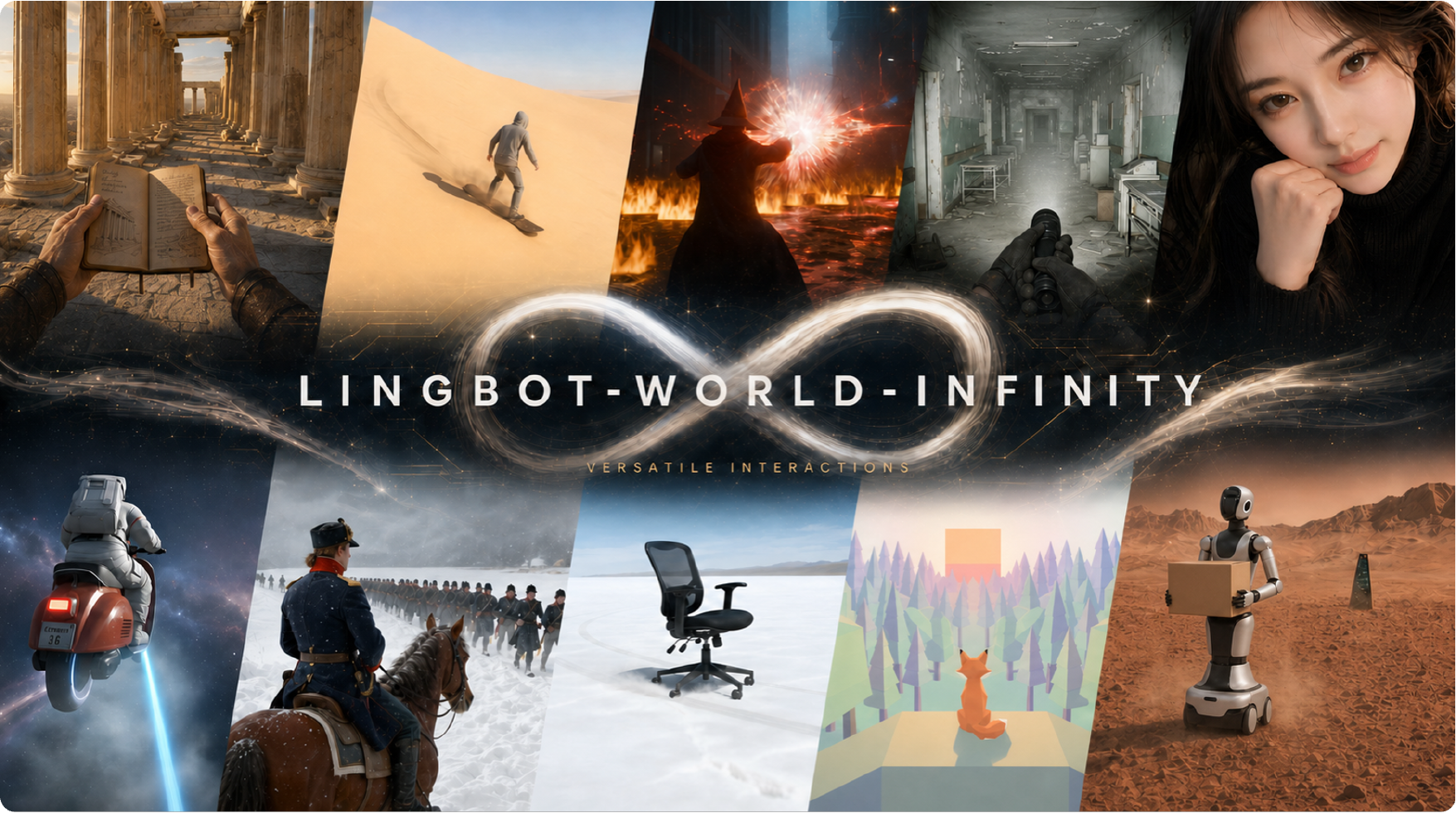}
\caption{\method generates infinite worlds in real time, featuring versatile interactions.}
\label{fig:teaser}
\end{figure}

\justifying
\section{Introduction}
Interactive world models~\cite{genie3,Gamegen-x,Yume,matrix2,matrix3,Worldplay,team2026dreamx}, generative systems that synthesize an environment frame by frame in response to a stream of user or agent actions, have recently emerged as a promising substrate for game generation ~\cite{Gamegen-x,solaris,matrix2,matrix3,worlddirector} and
embodied simulation ~\cite{cosmos,Abot-physworld,giga}. Driven by advances in causal (autoregressive) video generation ~\cite{CasualVid,MAGI-1,SkyReels,yang2025longlive,longlive_2.0,minWM},
such models can in principle render an explorable world that unfolds indefinitely and reacts to its inhabitants in real time. Yet turning this principle into a usable system has remained difficult, for two reasons.

The first concerns \textbf{long-horizon stability}. Because each frame is conditioned on the frames generated before it, errors are fed back into the model and accumulate; over time, textures smear, geometry warps, and the scene drifts away from any plausible world. Most existing systems therefore remain visually stable for only seconds to a few minutes before degrading, far short of the persistence one expects from a world meant to be lived in. The second concerns \textbf{interactivity at high fidelity}, which is computationally expensive: rendering detailed video while reacting to live input strains compute
budgets, and prior work has typically bought interactivity by sacrificing resolution, smoothness, or control, leaving the user with little more than coarse camera movement through an otherwise inert scene.

We argue that these two limitations are what stand between current models and genuinely open-ended worlds, and we tackle them directly. Our starting point is a causal video generation model trained to resist the error accumulation described above; the resulting backbone holds its quality far longer than prior open models and, in turn, makes a clean teacher for distillation. The distilled student is what makes the system practical: it sustains an explorable world that never has to end, at a resolution and frame rate high enough to feel responsive rather than merely playable. We stress-test this claim with an hour-long uninterrupted session in which quality shows no visible decay, evidence that the model's stability is structural rather than a short-lived artifact of a favorable clip. And rather than confining the user to wandering a frozen landscape, the model responds to an expressive vocabulary of actions such as combat, archery, spell-casting, and ranged fire, and even reshapes the environment itself, summoning weather such as snow or rain on demand.

Even a capable backbone, however, does not \emph{play itself}~\cite{Hu2024StoryAgentCS,Lian2023LLMgroundedVD,Huang2023FreeBloomZT}. Here we take a cue from how today's coding assistants are built: a strong language model becomes a working software engineer only once it is placed inside a harness, a scaffold such as Codex that lets it inspect state, act, and chase a goal across many turns. We adopt the same philosophy for world modeling. Around our generative core we wrap an \textbf{agentic harness} in which a \emph{pilot} agent reads the current scene and decides what the controllable character should do next, whether stepping forward, swinging at a foe, or reaching for a nearby object, while a \emph{director} agent keeps the world from running dry, seeding fresh content, props, and events as exploration proceeds. The two together turn a frame predictor into something that behaves like a world: self-sustaining, goal-directed, and open-ended, with no author scripting each moment in advance.

In summary, this paper makes the following contributions:

\begin{itemize}
    \item \textbf{An open, state-of-the-art causal world model} whose key advance is durability: leading visual quality paired with strong resistance to drift, giving the community both a usable backbone and a capable distillation teacher.

    \item \textbf{A real-time distilled model} that renders an unbounded, drift-free interactive world at 720p and 60\,fps, verified by over an hour of continuous generation without quality loss.

    \item \textbf{A rich, controllable action space} that goes well beyond navigation to include character actions such as combat, archery, spell-casting, and shooting, alongside on-the-fly environmental changes.

    \item \textbf{A pilot and director agentic harness} that orchestrates the model into a living interactive experience.
\end{itemize}

\begin{table}[t]
    \small
    \centering
    \caption{
        \textbf{Comparison with recent interactive world models.} \method stands out as the only model to achieve hour-level (infinite) generation duration within a general domain. Furthermore, it uniquely combines this extended continuous generation with high dynamic degree, semantic interaction, and real-time performance, while being fully open-sourced.
    }
    \label{tab:comparison}
    \SetTblrInner{rowsep=1.2pt}      
    \SetTblrInner{colsep=4.6pt}     
    \definecolor{linegray}{HTML}{BDBDBD} 
    \definecolor{bg_gray1}{HTML}{FAFAFA}
    \definecolor{bg_gray2}{HTML}{F2F2F2}
    \definecolor{bg_gray3}{HTML}{EAEAEA}
    \definecolor{bg_gray4}{HTML}{E2E2E2}
    \definecolor{bg_gray5}{HTML}{DADADA}
    \definecolor{bg_purple}{HTML}{6A67F3}
    \begin{tblr}{
        cells={halign=l,valign=m},
        column{1}={bg=white},
        column{2}={bg=bg_gray1},
        column{3}={bg=bg_gray2},
        column{4}={bg=bg_gray3},
        column{5}={bg=bg_gray4},
        column{6}={bg=bg_gray5},
        column{7}={bg=bg_purple, fg=white},
        hline{2}={0.5pt, fg=linegray},
    }
    \ & \textbf{M-G 3.0}~\cite{matrix3} & \textbf{D-W}~\cite{team2026dreamx} & \textbf{LingBot-World}~\cite{lingbot-world} & \textbf{HappyOyster}~\cite{happyoyster} & \textbf{Genie 3}~\cite{genie3} & \textbf{Ours} \\
    Generation Duration & Minutes & Minutes & Minutes & Minutes & Minutes & Hours (Infinite) \\ 
    Semantic Interaction & None & None & None & Few & Few & Infinite \\
    Domain & Game & General & General & General & General & General \\ 
    Dynamic Degree & Medium & Medium & High & Medium & Medium & High \\ 
    Real-time & \ding{51} & \ding{51} & \ding{51} & \ding{51} & \ding{51} & \ding{51} \\ 
    Open-source & \ding{51} & \ding{51} & \ding{51} & \ding{55} & \ding{55} & \ding{51} \\ 
    \end{tblr}
\end{table}
\section{Data pipeline} \label{sec:data}

To train an interactive world model that remains responsive to time-varying instructions and control signals, we construct a unified data engine that transforms heterogeneous raw videos into temporally localized training dataset. The engine consists of three stages: \textbf{data acquisition}, \textbf{data profiling}, and \textbf{multi-granularity annotation}. 
In the first stage, videos from different sources are normalized into a shared metadata schema. In the second stage, low-quality or semantically unsuitable samples are removed through technical filtering and VLM-based profiling. 
In the final stage, retained videos are converted into structured annotations, video-level global captions, and chunk-wise local captions. The design goal is to provide both global semantic context and local supervision for state updates. Each chunk combines reliable visual evidence with concise local-state descriptions, helping the model capture scene dynamics, interaction outcomes, and control-sensitive transitions.

\subsection{Data Acquisition}

Our corpus is assembled from three complementary sources: self-collected egocentric videos, synthetic data from games and Unreal Engine environments, and large-scale web videos~\cite{yang2026aoe,wang2025spatialvid,grauman2022ego4d,buildaiegocentric10k2025,damen2020epic,egodex,bansal2022my,jang2019epic}. Egocentric videos capture real-world first-person interactions and natural hand-object behaviors. Synthetic data provides accurate scene geometry and temporally aligned interaction signals, such as jumping, attacking, driving and flying. These signals are difficult to recover reliably from in-the-wild videos, but are important for learning controllable state transitions. Web videos provide scalable open-domain coverage and expose the model to long-tail visual distributions.

Before profiling, each raw video is converted into a standardized record containing source information and basic attributes for downstream processing. This mixed-source data curation strategy follows recent video world model data pipelines, which combine web-scale, synthetic, and action-conditioned data to improve both visual diversity and controllability~\cite{xiong2026actworld,lingbot-world,cosmos}.

\subsection{Data Profiling}

Raw videos differ in quality, viewpoint, and semantic content, so we first validate decoding, remove invalid samples, and detect shot boundaries. For long videos, we use TransNet V2~\cite{soucek2024transnet} to split them into temporally coherent clips, which are then filtered by basic constraints such as duration, resolution, and decoding stability.
After segmentation, we apply technical scoring filters as an inexpensive screening step. The score aggregates signals such as visual quality, luminance range, OCR-based text occupancy, motion statistics, and encoding stability~\cite{schuhmann2022aesthetic,wu2023exploring,cui2025paddleocr}. This stage is designed to remove clearly unsuitable clips while preserving enough high-quality candidates for later semantic filtering.

We then use a vision-language model to profile the retained candidate clips~\cite{qwen3.5,qwen3.6-35b-a3b,qwen3.6-27b}. VLM profiling produces two groups of attributes. The first group describes sample quality, including editing artifacts, and ambiguous visual evidence. The second group describes semantic content, including viewpoint, activity category, and interaction pattern. Technical filtering and VLM profiling serve different purposes, the former provides low-cost quality control, while the latter organizes the retained clips for data balancing and prompt routing.

\begin{figure}[t]
    \centering
    \includegraphics[width=\linewidth]{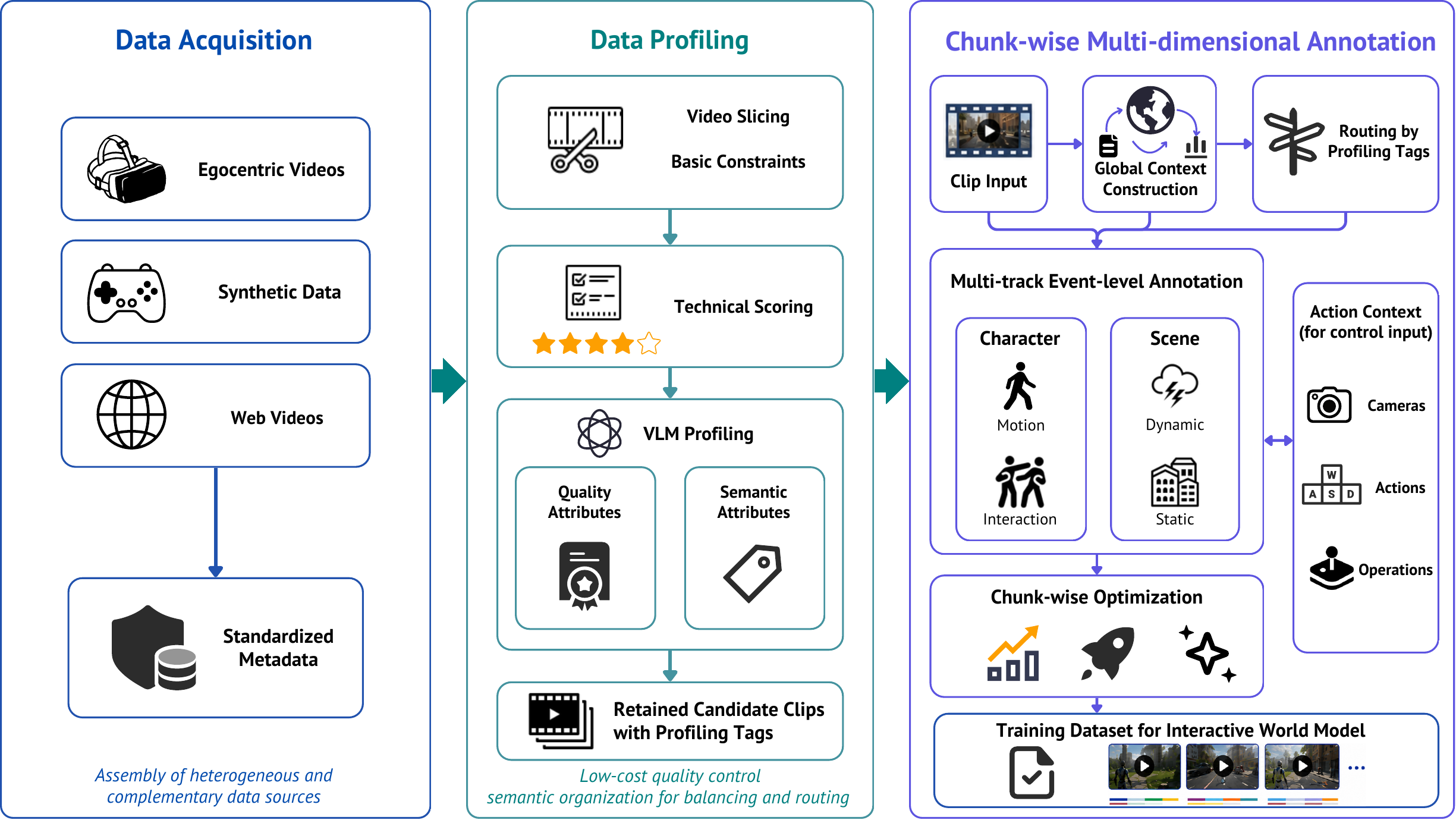}
    \caption{
    Overview of the proposed data engine. Heterogeneous raw videos are temporally segmented, filtered, and routed to category-specific annotation pipelines, producing optimized chunk-wise captions.
    }
    \label{fig:data_engine}
\end{figure}

\subsection{Multi-dimensional Video Annotation}

Existing video datasets often associate each video with a single global caption, which provides a video-level semantic summary of the scene, task context, and overall interaction trajectory~\cite{chen2026memdreamerdecouplingperceptionreasoning, zhang2026muss, wang2025koala, chen2024panda, tan2024vidgen, nan2025openvid, Yang2025ViMix14MAC}.
However, our model is designed for real-time interaction, where the conditioning signal may vary over time in response to user instructions or actions. 
Training only with global captions would therefore create a train--inference mismatch: the model would be required to follow localized instructions at inference time without having been exposed to comparable temporal conditioning during training.

To address this issue, we construct multi-dimensional annotations that combine video-level global context with temporally localized chunk-wise descriptions. 
Given a preprocessed video clip, our pipeline first establishes a stable global context, then produces event-level annotations for each temporal chunk, and finally composes these annotations into visually grounded and temporally aligned captions. This hierarchical design is consistent with recent interactive and causal video world models that leverage hierarchical captions or per-chunk semantic descriptions~\cite{meng2026causalcine,longlive_2.0,Yesiltepe_2026_CVPR}.

\paragraph{Multi-track event-level annotation} 
For each chunk, the annotation process is adapted according to the clip profile and available input modalities.
Rather than enforcing a single temporal boundary for all semantic attributes, we annotate multiple event tracks independently. These tracks cover subject visibility, motion state, interaction state, environmental dynamics, and static scene state. 
This decoupled representation reduces temporal ambiguity by preventing unrelated visual, interaction, and scene changes from being merged into a single description. 
When control signals are available, we use temporally aligned and semantically normalized action context as auxiliary evidence for chunk annotation. 
The action context mainly provides boundary cues and helps recover short interactions that may be missed by sparse visual sampling, while the final captions remain constrained by visible evidence to avoid speculative actions or state changes.

\paragraph{Chunk-wise optimization}
After event-level annotation, we compose active track states into chunk-wise captions and apply a lightweight refinement stage to improve terminology consistency and temporal smoothness. 
We further remove future-revealing, speculative, or redundant expressions, yielding visually grounded and temporally aligned chunk-wise annotations for training.
\section{LingBot-World-Infinity}\label{sec:method}

\begin{figure*}[t]
\centering
\includegraphics[width=\linewidth]{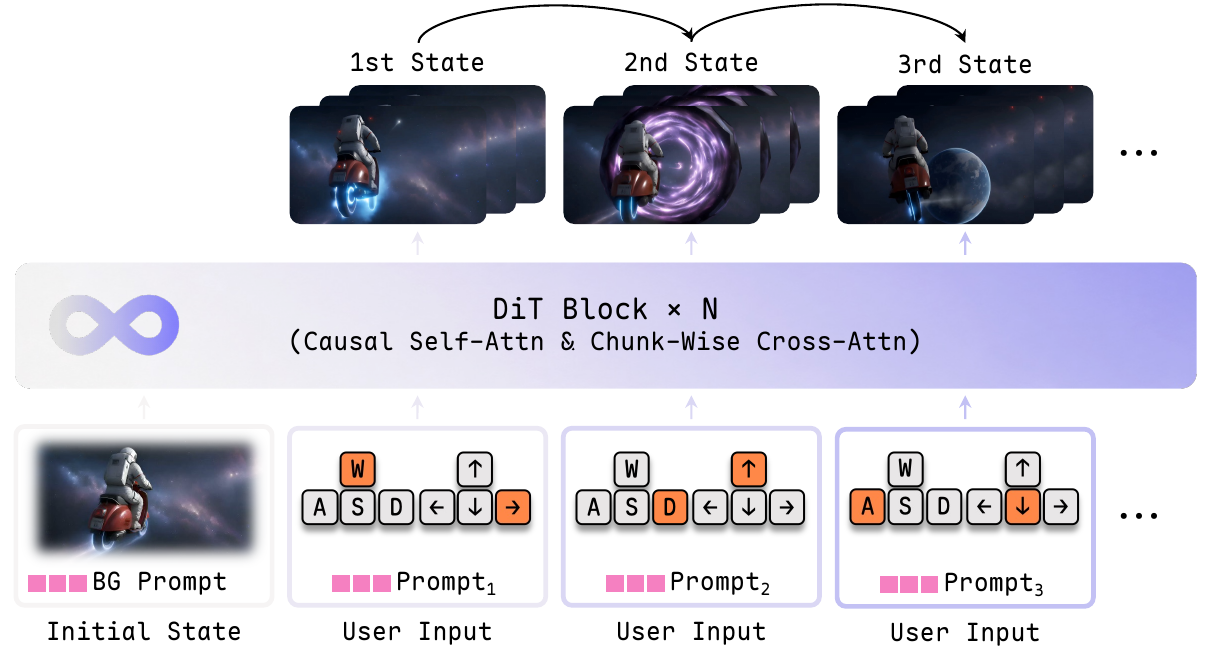}
\caption{\textbf{Overview of \method Pipeline.} 
An interactive world simulator is implemented as a causal video model.
Our Infinity World is initialized from an initial image and its background description.
The future world states are then autoregressively generated, conditioned on the historical context and user inputs (camera poses and prompts).}
\label{fig:model_pipeline}
\end{figure*}

As shown in \cref{fig:model_pipeline}, {\method} takes an initial frame and interacts with a stream of user inputs to autoregressively generate an infinite world that extends endlessly in real time and remains drift-free over arbitrarily long horizons.
To this end, we first formalize the interactive world simulator as a causal generative process (\cref{sec:formulation}). 
Building on this formulation, we adopt a two-stage strategy to train {\method}: a \textbf{pre-training} stage that learns a causal, action-conditioned world model~(\cref{sec:pretraining}), and a \textbf{post-training} stage that distills the model into a real-time generator and suppresses long-horizon drift~(\cref{sec:posttraining}).

\subsection{Formulation}
\label{sec:formulation}
\begin{quote}
\itshape ``The cause must be prior to the effect.'' \par
\normalfont\hfill --- David Hume, \emph{A Treatise of Human Nature}~\cite{hume2000treatise}
\end{quote}
The first principle of causal reasoning is that an effect never precedes its cause: the future is shaped by the past, never the reverse.
We follow this principle to introduce {\method}, an interactive world simulator where each state is generated from past visual observations and the current user input.
We therefore cast world simulation as a causal generative process along the time axis. 
Let $\mathcal{V} = \{x_1, x_2, \dots, x_T\}$ denote a sequence of video frames, where $x_t \in \mathbb{R}^{H \times W \times C}$ represents the visual state at time step $t$, and let $\mathcal{A} = \{a_1, a_2, \dots, a_T\}$ denote the corresponding sequence of user inputs (including control signals and action prompts).
Under the causal assumption, each state depends solely on its historical context $(x_{<t}, a_{\le t})$, yielding the factorization:
\begin{equation}
\label{equ:causal_factorization}
    p_\theta(x_{1:T} \mid a_{1:T}) = \prod_{t} p_\theta\!\left(x_t \mid x_{<t}, a_{\le t}\right)
\end{equation}
where $\theta$ parameterizes the world transitions and dynamics that {\method} learns by maximizing this causal likelihood over large-scale video data.
In this work, we enforce causality natively throughout training, as detailed in the following two stages.

\subsection{Pre-Training: Causal World Model}
\label{sec:pretraining}
\begin{figure*}[t]
\centering
\includegraphics[width=\linewidth]{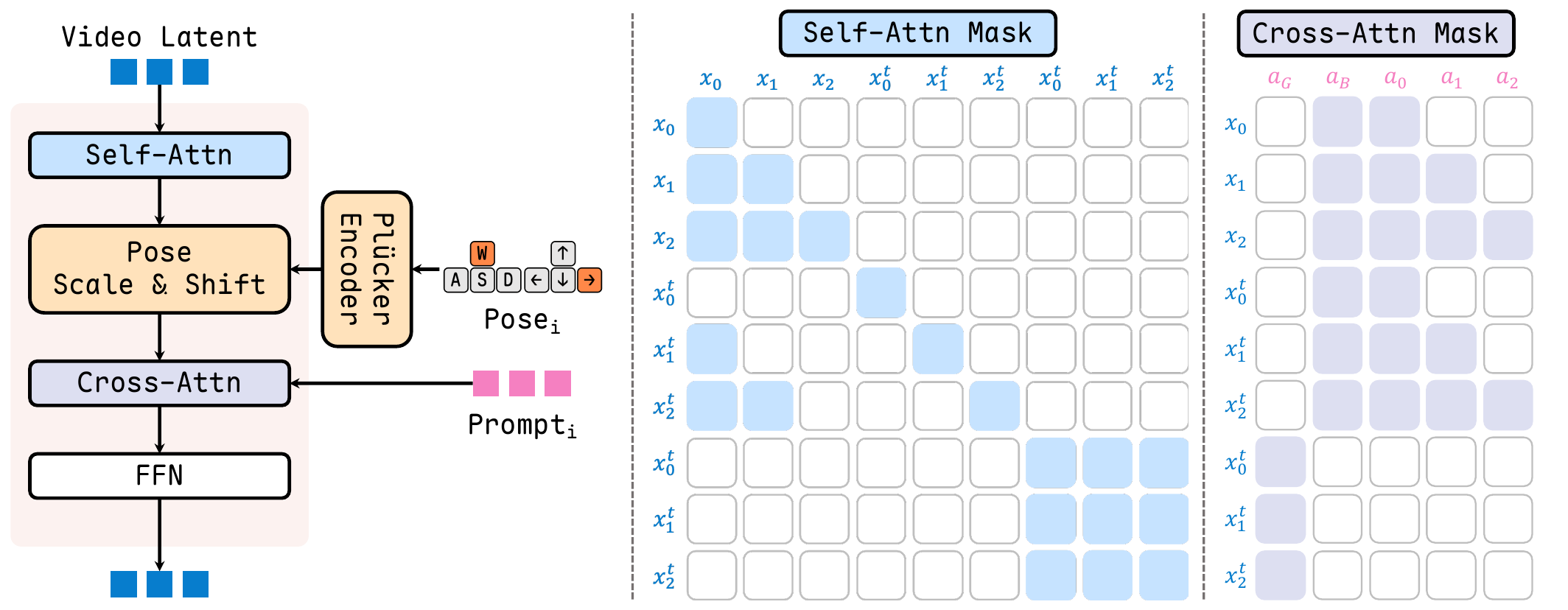}
\caption{\textbf{Overview of \method DiT Block and MoBA Attention Mask.} 
The action comprises camera poses and chunk-wise prompts, injected into the DiT block to enable user interaction.
For self-attention, a bidirectional block is appended to teacher forcing mask, enabling autoregressive generation while preserving visual fidelity.
For cross-attention, the autoregressive component attends to a background prompt and chunk-wise prompts of lower-triangular pattern to prevent access to future information, while the bidirectional component attends to a global prompt.}
\label{fig:attn_mask}
\end{figure*}

In the pre-training stage, we train a causal world model that generates a boundless, action-controllable video world with high visual fidelity.
As illustrated in \cref{fig:attn_mask}, we support two forms of user input, camera poses and prompts, as actions, enabling rich interactive control.
Moreover, we propose \textbf{Mixture of Bidirectional and Autoregressive Attention Mask}, a causal self-attention mask where a bidirectional component is appended to the teacher forcing~\cite{williams1989learning} mask, enabling autoregressive generation while preserving the fidelity of the base generator.
We also design a causal cross-attention mask for leak-free conditioning and flexible interaction.

\noindent\textbf{Action Representation.}
In {\method}, the action comprises two forms of user input: camera poses and textual prompts.
For camera control, we represent camera pose using {Pl\"ucker embeddings}~\cite{he2024cameractrl,he2025cameractrl}, which encode the viewing ray of each pixel as six-dimensional coordinates, and utilize an adaptive layer normalization~(AdaLN) mechanism~\cite{xu2019understanding} to incorporate these control signals into the diffusion process without disrupting the pre-trained visual priors.
For textual control, we adopt chunk-wise prompts for autoregressive video generation, where each video chunk is assigned its own caption, enabling time-localized semantic control. We detail how these chunk-wise prompts are injected via cross-attention mask below.

\noindent\textbf{Mixture of Bidirectional and Autoregressive~(MoBA) Attention Mask.}
With teacher forcing mask, the model predicts the current state from the clean context for autoregressive video generation.
However, we find that the longer the context grows, the more our model learns to rely solely on the context instead of predicting future frames, resulting in overfitting and visual quality degradation.
Therefore, we propose MoBA mask, a hybrid attention mechanism, to counter this issue by integrating a bidirectional component into teacher forcing mask.
As illustrated in \cref{fig:attn_mask}, the self-attention MoBA mask starts from teacher forcing, where each noisy frame $x_i^{t}$ (frame $i$ at noise timestep $t$) attends only to itself and its clean context $x_{<i}$.
A bidirectional component with full attention (the bottom-right block of the mask) is then integrated, helping the model adapt to flexible-length video generation and promoting its transition from bidirectional to autoregressive generation. 
It also serves as a regularizer that mitigates the overfitting of pure teacher forcing, thereby avoiding visual quality degradation.
To match two components of MoBA, we design a corresponding cross-attention mask. 
For the teacher forcing component, each frame attends to a background prompt $a_{B}$ together with the chunk-wise prompts $a_{\le i}$ of the current and all preceding chunks in a lower-triangular pattern, preventing any future semantics from leaking in. 
For the bidirectional component, which already attends over all frames, we follow standard practice and use a global prompt $a_{G}$ that describes all events throughout the video.

\noindent\textbf{Training objective.}
We optimize {\method} with a conditional flow-matching objective. 
For each frame $i$, we construct a noisy latent $x_i^{t} = (1-t)\,x_i + t\,\epsilon$ with $t \sim \mathcal{U}(0,1)$ and $\epsilon \sim \mathcal{N}(0,\mathbf{I})$, and train the network $v_\theta$ to predict the flow velocity conditioned on the clean context $x_{<i}$, the camera pose $p_{\le i}$, and the prompt $a_{\le i}$:
\begin{equation}
\label{equ:pretrain_loss}
    \Loss_{\text{fm}} = \E_{x,\, i,\, t,\, \epsilon}
    \left\| v_\theta\!\left(x_i^{t},\, t \mid x_{<i},\, p_{\le i},\, a_{\le i}\right) - (\epsilon - x_i) \right\|^2,
\end{equation}
where $p_{\le i}$ and $a_{\le i}$ denote the camera poses and prompts up to the current chunk, and the target velocity $\epsilon - x_i$ follows the rectified-flow interpolation.

\subsection{Post-Training: Few-Step Distillation}
\label{sec:posttraining}
In the post-training stage, we compress the multi-step pre-trained causal diffusion world model into a few-step generator suitable for real-time interaction, while mitigating the error accumulation that arises over long autoregressive rollouts.
To this end, we combine \emph{consistency distillation}~\cite{song2023consistency} to reduce the number of denoising steps with \emph{distribution matching distillation} (DMD)~\cite{yin2024dmd} to improve fidelity and suppress rollout drift.

\noindent\textbf{Consistency Distillation.}
Although the pre-trained teacher produces high-quality frames, generating each frame requires many denoising steps, which is prohibitively expensive for interactive use.
We therefore distill the teacher into a consistency model~\cite{song2023consistency, lin2025apt2, zheng2026rcm, zhu2026causal} $G_\theta$, which predicts a trajectory-invariant target from an intermediate noisy latent.
In particular, latent states lying on the same teacher probability-flow ODE (PF-ODE) trajectory should map to identical predictions under $G_\theta$.
Let $x_i^t$ denote the latent state of frame $i$ at diffusion time $t$, and let $\tilde{x}_i^{\,t-\Delta t}$ denote the latent obtained by integrating the teacher PF-ODE from $(x_i^t, t)$ to time $t-\Delta t$, where $\Delta t > 0$.
Under the causal conditioning $c = (x_{<i},\, p_{\le i},\, a_{\le i}),$ we enforce local consistency between adjacent points on the same teacher trajectory by minimizing
\begin{equation}
\label{equ:cd_loss}
    \Loss_{\mathrm{CD}}
    =
    \E\!\left[
    d\!\left(
    G_\theta(x_i^t, t \mid c),\;
    G_{\theta^-}(\tilde{x}_i^{\,t-\Delta t}, t-\Delta t \mid c)
    \right)
    \right],
\end{equation}
where $d(\cdot,\cdot)$ is a distance metric and $\theta^-$ denotes an exponential moving average (EMA) of the student parameters $\theta$.
The expectation is taken over training samples, diffusion times, and trajectory points induced by the teacher dynamics. 
Optimizing~\cref{equ:cd_loss} distills the teacher's multi-step denoising trajectory into a few-step student generator while preserving the action-conditioned dynamics acquired during pre-training.

\noindent\textbf{Distribution Matching Distillation.}
While consistency distillation substantially reduces the sampling cost, the resulting few-step student generator may still suffer from degraded visual fidelity and compounding drift during long-horizon self-rollout.
To further refine the student, we employ distribution matching distillation (DMD)~\cite{yin2024dmd}, which optimizes the generator using the gradient of the KL divergence between the noised student distribution and the noised data distribution.
Let $\hat{x}_i$ denote a sample generated by the student, and let $\hat{x}_i^t$ be its noised version obtained by forward diffusion to time $t$.
Then the gradient of the objective is given by
\begin{equation}
\label{equ:dmd_loss}
    \nabla_\theta \E\left[D_{\mathrm{KL}}\left(p_{\theta,t} \,\|\, p_{\mathrm{data},t}\right)\right]
    =
    - \E\!\left[
    \left(
    s_{\mathrm{real}}(\hat{x}_i^t, t \mid c)
    -
    s_{\mathrm{fake}}(\hat{x}_i^t, t \mid c)
    \right)
    \frac{\partial \hat{x}_i}{\partial \theta}
    \right],
\end{equation}
where the conditioning variable $c = (x_{<i},\, p_{\le i},\, a_{\le i})$ is defined as in~\cref{equ:pretrain_loss}.
Here, $s_{\mathrm{real}}$ and $s_{\mathrm{fake}}$ approximate the scores of the noised data distribution and the current noised student distribution, respectively. The expectation is taken over sampled diffusion times and student-generated samples.
To improve robustness under deployment-time dynamics, we apply DMD over long self-rollout trajectories rather than only on teacher-forced states~\cite{huang2025self}. 
As a result, the student is optimized on the state distribution induced by its own predictions, which reduces accumulated drift over extended autoregressive rollouts.
\section{Deployment}\label{sec:eval}

Our goal is real-time interactive video generation with low latency, rich interaction capabilities, high visual quality, and user-friendly interface. To this end, we develop an inference stack consisting of four components: systematic optimization, agentic interaction harness, visual quality enhancement, and user interface design. The following sections describe each component.

\begin{figure*}[t]
\centering
\includegraphics[width=0.85\linewidth]{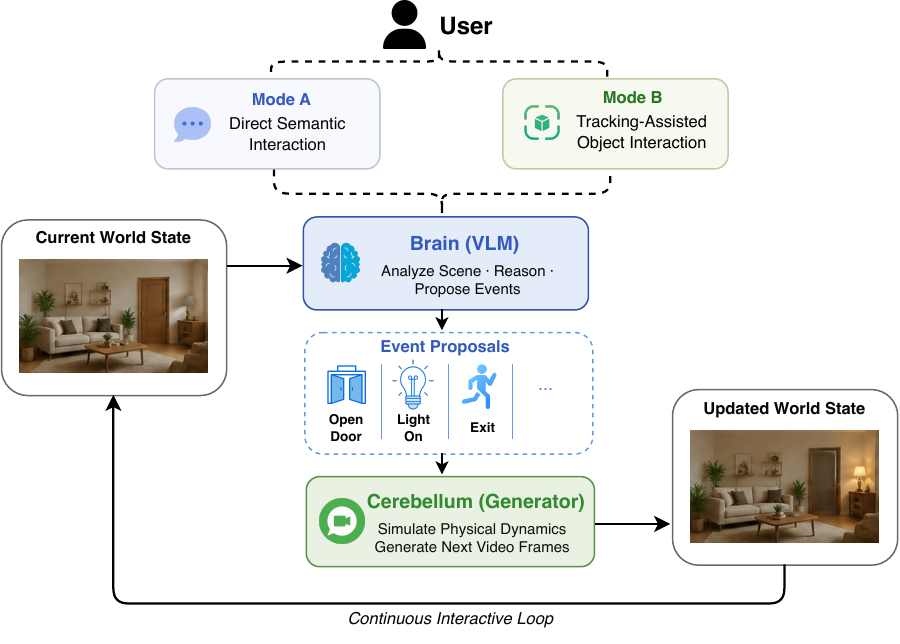}
\caption{Overview of the Agentic Interaction Harness. Users can either interact with the existing world through semantic or object-centric actions, or intervene by introducing high-level textual events. The VLM (Director) performs causal reasoning and proposes coherent event updates, while the Video Generator (Pilot) grounds these semantic decisions into physically consistent video rollouts, enabling continuous interactive world simulation.}
\label{fig:harness}
\end{figure*}

\begin{figure*}[t]
\centering
\includegraphics[width=\linewidth]{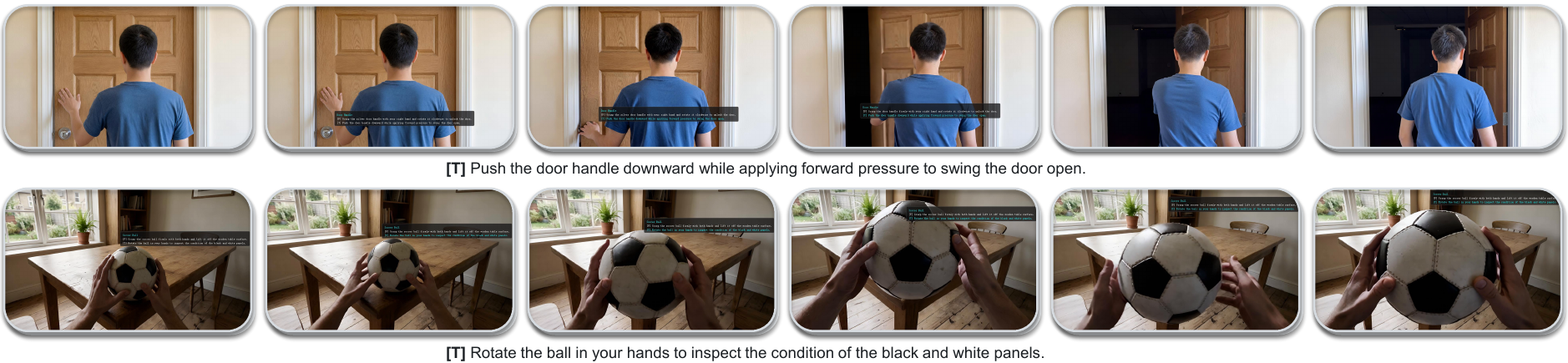}
\caption{In the tracking-mode interface, the Vision-Language Model (VLM) comprehends interactive objects within the scene, while the tracking model continuously tracks these targets to display dynamic interactive floating windows (event cards) in real-time. Powered by the "Director-Pilot" co-simulation framework, the model demonstrates robust interactive capabilities by performing causal reasoning based on user actions (e.g., pushing a door open or rotating a soccer ball) and rendering physically logical, highly coherent spatio-temporal dynamics.}
\label{fig:sam_interactive}
\end{figure*}

\begin{figure*}[!ht]
\centering
\includegraphics[width=\linewidth]{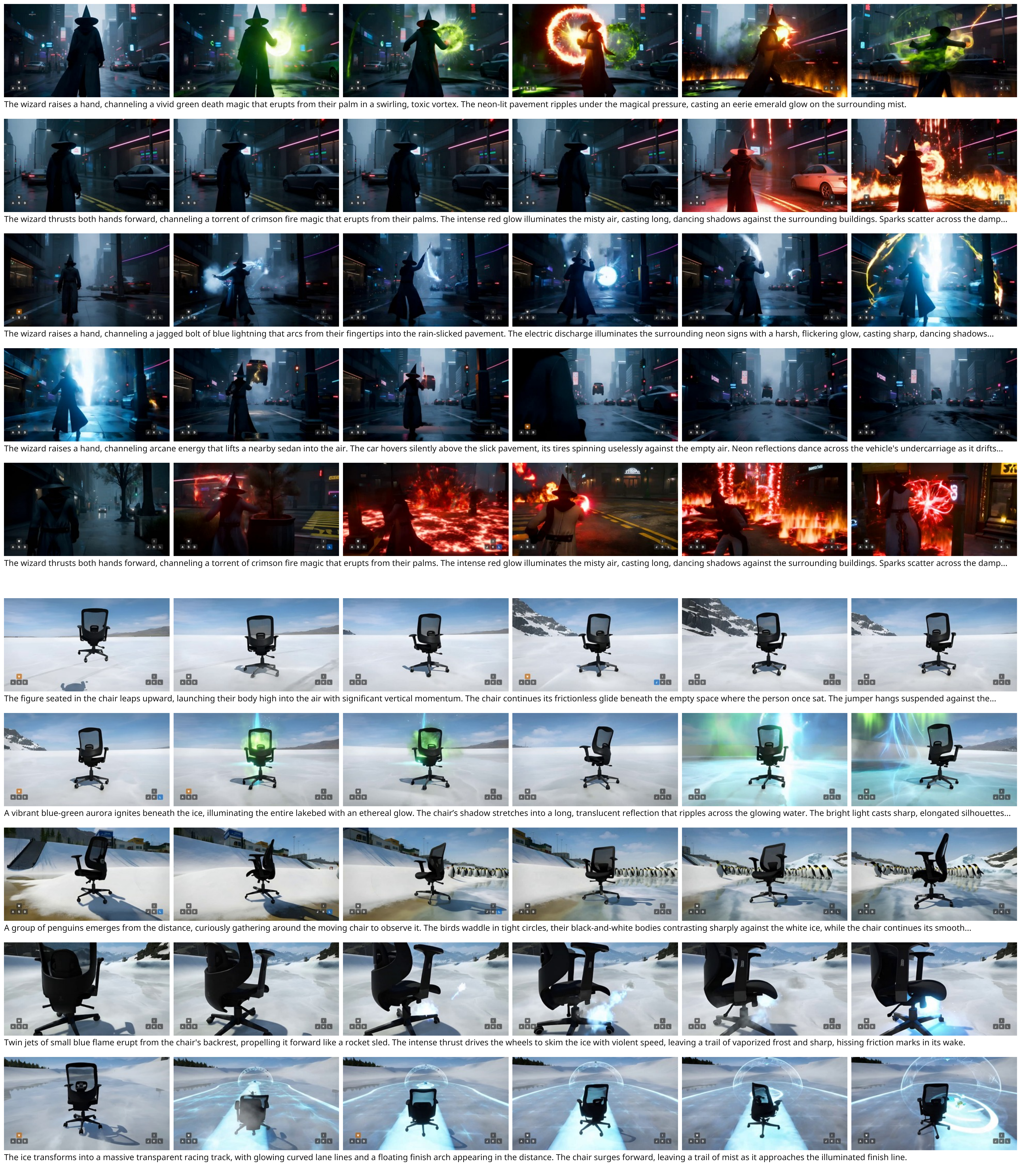}
\caption{Our world model enables controllable world exploration with versatile interactions. It supports flexible prompt switching across different world horizons, allowing the scene semantics to evolve smoothly, while also enabling controllable navigation of diverse protagonists and objects throughout the generated world (1/2).}
\label{fig:result1}
\end{figure*}

\begin{figure*}[!ht]
\centering
\includegraphics[width=\linewidth]{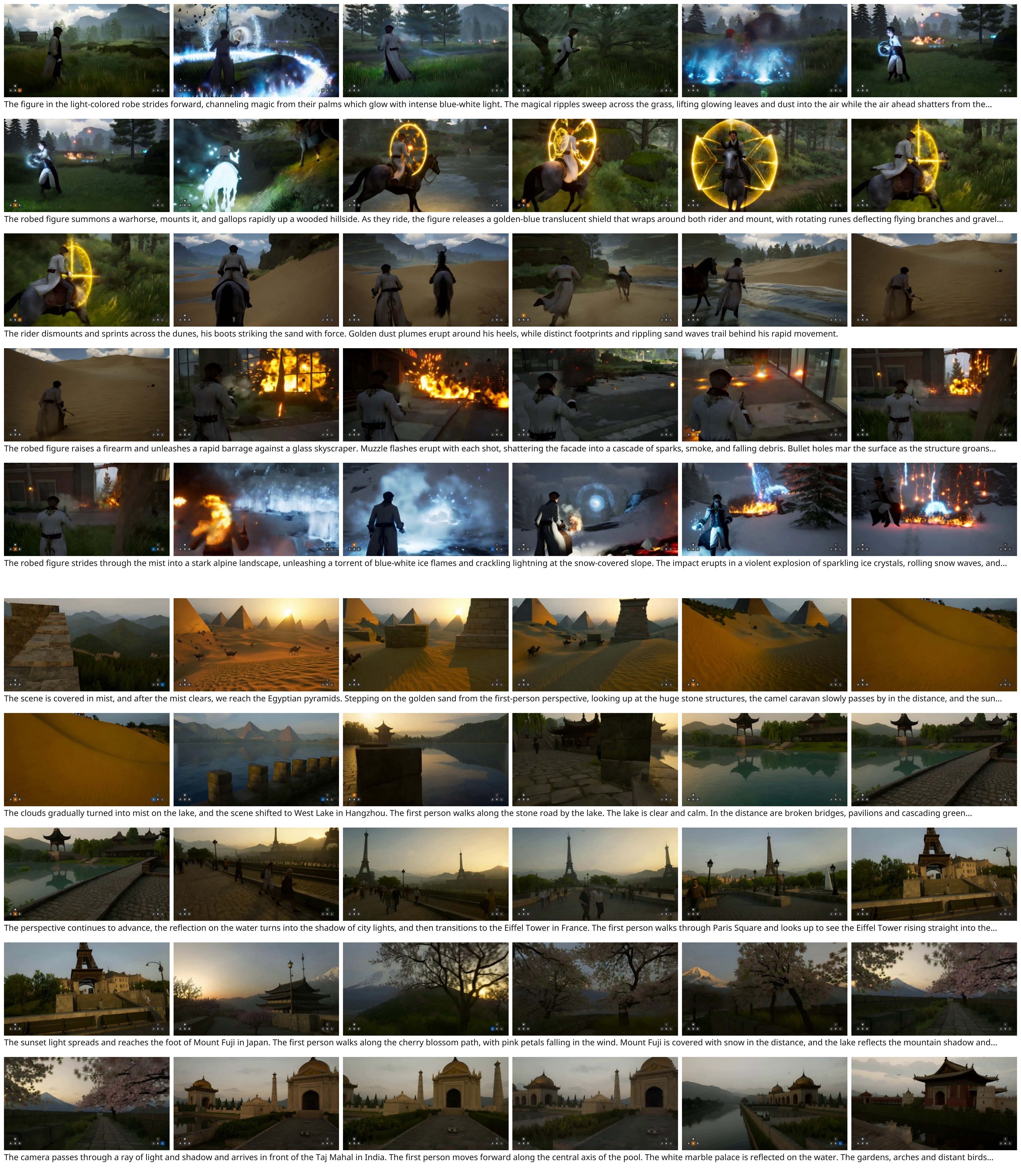}
\caption{Our world model enables controllable world exploration with versatile interactions. It supports flexible prompt switching across different world horizons, allowing the scene semantics to evolve smoothly, while also enabling controllable navigation of diverse protagonists and objects throughout the generated world (2/2).}
\label{fig:result2}
\end{figure*}

\subsection{Systematic Optimization}

Inference efficiency is crucial for enabling interactive video generation ~\cite{hacohen2024ltx, Fang2024xDiTAI, feng2025streamdiffusionv2, flashdreams2026}. In our deployment setting, inference efficiency is required along two complementary dimensions. The first is \emph{model inference efficiency}, which improves the throughput of the video generation backbone to approach real-time generation. The second is \emph{streaming responsiveness}, which reduces the end-to-end latency from user input to the first visible response, thereby improving the perceived interactivity of the system.

\noindent\textbf{Model inference efficiency.}
To improve generation throughput, we apply a collection of system-level optimizations to the distilled DiT backbone. We first leverage compiler-based optimization together with efficient attention kernels to reduce runtime overhead and accelerate the iterative denoising process. We further adopt a hybrid parallel inference strategy that distributes computation across multiple GPUs while maintaining efficient communication, enabling scalable inference for long video sequences. To avoid downstream bottlenecks, we pipeline latent generation and frame reconstruction asynchronously. Latent generation continues without waiting for image decoding, while VAE decoding is performed on dedicated workers with parallelism. This design improves hardware utilization and shortens the critical path of video generation.

\noindent\textbf{Streaming responsiveness.}
Beyond improving raw generation throughput, we optimize the system to minimize perceived interaction latency. Instead of waiting for an entire video chunk to finish decoding, decoded results are streamed incrementally as they become available, allowing decoding, transmission, and subsequent generation to overlap. We further employ an efficient video streaming pipeline for low-latency delivery to the client, reducing the time to the first visible frame.

\subsection{Agentic Interaction Harness}
\label{subsec:interactive_event_proposal}

While standard video generation models excel at synthesizing short-term visual dynamics, they inherently lack the causal reasoning required for open-ended, logical world simulation. To bridge the gap between passive video generation and interactive world modeling, we introduce a novel \textbf{Director-Pilot Co-Simulation Framework}. In this architecture, a Vision-Language Model (VLM) serves as the ``Director'' of the world, governing macroscopic semantic rules and causal reasoning, while the underlying Video Generator (Diffusion Transformer) acts as the ``Pilot'', responsible for simulating low-level physical dynamics and rendering high-fidelity visual transitions, as is shown in Fig. \ref{fig:harness}. 
Through this dual-system design, the VLM continuously analyzes the current visual phenomena, anticipates the logical consequences of user interactions, and formulates explicit ``event proposals''. These proposals are subsequently fed into the Video Model, which grounds the semantic instructions into coherent spatio-temporal rollouts. This framework empowers users to interact with the simulated world through two primary modalities: Agentic Interaction Reasoning and Prompt-Driven World Intervention.

\subsubsection{Agentic World Interaction}
In the interactive simulation, users can perform actions within the environment, and the ``Director'' (VLM) is tasked with predicting the outcomes of these behaviors based on physical and semantic common sense. To accommodate different levels of interaction granularity, we design two distinct interaction modes:

\begin{itemize}
    \item \textbf{Mode A: Direct Semantic Interaction.} In this default mode, the VLM directly analyzes the current frame and generates dynamic event cards for the subject (e.g., the user's avatar). When a user triggers an interaction, the VLM evaluates the scene context and proposes immediate, logical consequences without requiring explicit object masks. This allows for seamless, holistic interactions with the environment.
    
    \item \textbf{Mode B: Tracking-Assisted Object Interaction.} For precise, object-centric manipulation, we integrate a SAM-based (Segment Anything Model) action-proposal loop, as in~\cref{fig:sam_interactive}. The VLM identifies specific interactive elements (e.g., non-human objects) within the scene and assigns targeted action proposals. SAM continuously tracks these objects across video chunks, maintaining spatial consistency. Users can select a tracked object and trigger specific actions (e.g., ``hold-to-switch'' mechanics). The VLM then deduces the state change of the tracked object, passing the updated semantic condition to the Video Model for precise dynamic execution.
\end{itemize}

\subsubsection{Agentic World Intervention}
Beyond physical interactions, our framework allows users to act as ``creators'' by directly intervening in the world state via textual prompts. Users can propose arbitrary events to alter the trajectory of the simulation. The VLM agent processes these interventions to ensure they are integrated logically into the ongoing narrative. This capability supports:

\begin{itemize}
    \item \textbf{Global State Shifts:} Users can dynamically alter macroscopic environmental factors, such as transitioning the time of day (e.g., day to night), changing weather conditions (e.g., summoning a snowstorm), or initiating global events (e.g., a festive celebration).
    
    \item \textbf{Local Entity Injection:} Users can command the sudden appearance of specific entities (e.g., spawning a flock of birds or a group of specific creatures). The VLM determines the most plausible spatial and temporal entry points for these entities, while the Video Model seamlessly renders their integration into the existing physical space.
\end{itemize}

\subsection{Visual Quality Enhancement}

\subsubsection{Spatio-temporal Refiner}
To improve visual quality, we append a lightweight spatio-temporal refiner after remote VAE decoding. The refiner first performs spatial refinement, upsampling the decoded frames and restoring sharper local details. It then performs temporal refinement, synthesizing intermediate frames to produce smoother motion and a higher frame rate. Both stages are compiled into TensorRT engines and executed with multi-GPU parallelism. The refined frames are delivered through the same asynchronous streaming pipeline, improving perceptual quality while introducing only limited additional latency to the whole pipeline.

\subsubsection{Dynamic KV Cache Management}
A key component of our real-time system is a dynamic KV-cache scheduling mechanism. Rather than maintaining a fixed context for every inference chunk, we adapt the cache on the fly according to the current control signal and input state. By conditioning the cache on this context, we retain the history that is most informative for the current chunk while discarding entries that contribute little. This selective scheduling yields two benefits at once. First, it reduces the effective context the model must attend to, which accelerates inference and helps sustain real-time throughput. Second, by focusing attention on the most relevant history, it suppresses interference from stale or unrelated content and thereby improves the quality and coherence of the generated frames.

\begin{figure}[t]
    \centering
    \includegraphics[width=\linewidth]{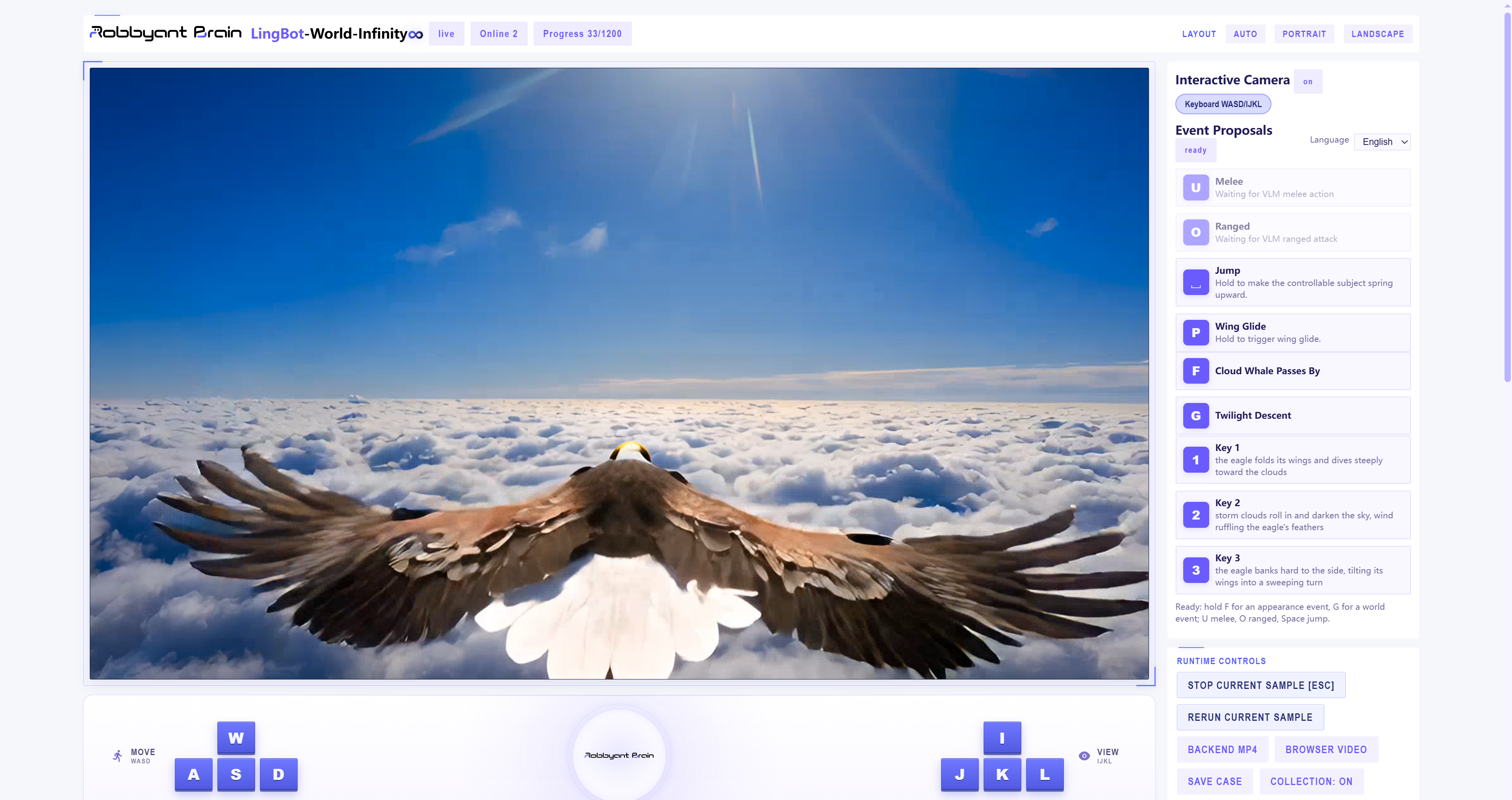}
    \caption{Interface of our interactive system. The center shows the live generation viewport; the bottom panel exposes the low-level control scheme (WASD for movement, IJKL for view control); and the right-hand panel is the agentic control surface, which lists the VLM-proposed ``Event Proposals,'' each bound to a hotkey. Fixed keys (\emph{Space}, \emph{P}) are always available, \emph{U}/\emph{O} and \emph{F}/\emph{G} carry context-aware
    character actions and environmental events proposed by the VLM, and the numeric keys are user-registered event slots.}
    \label{fig:ui}
\end{figure}

\begin{figure*}[ht]
\centering
\includegraphics[width=\linewidth]{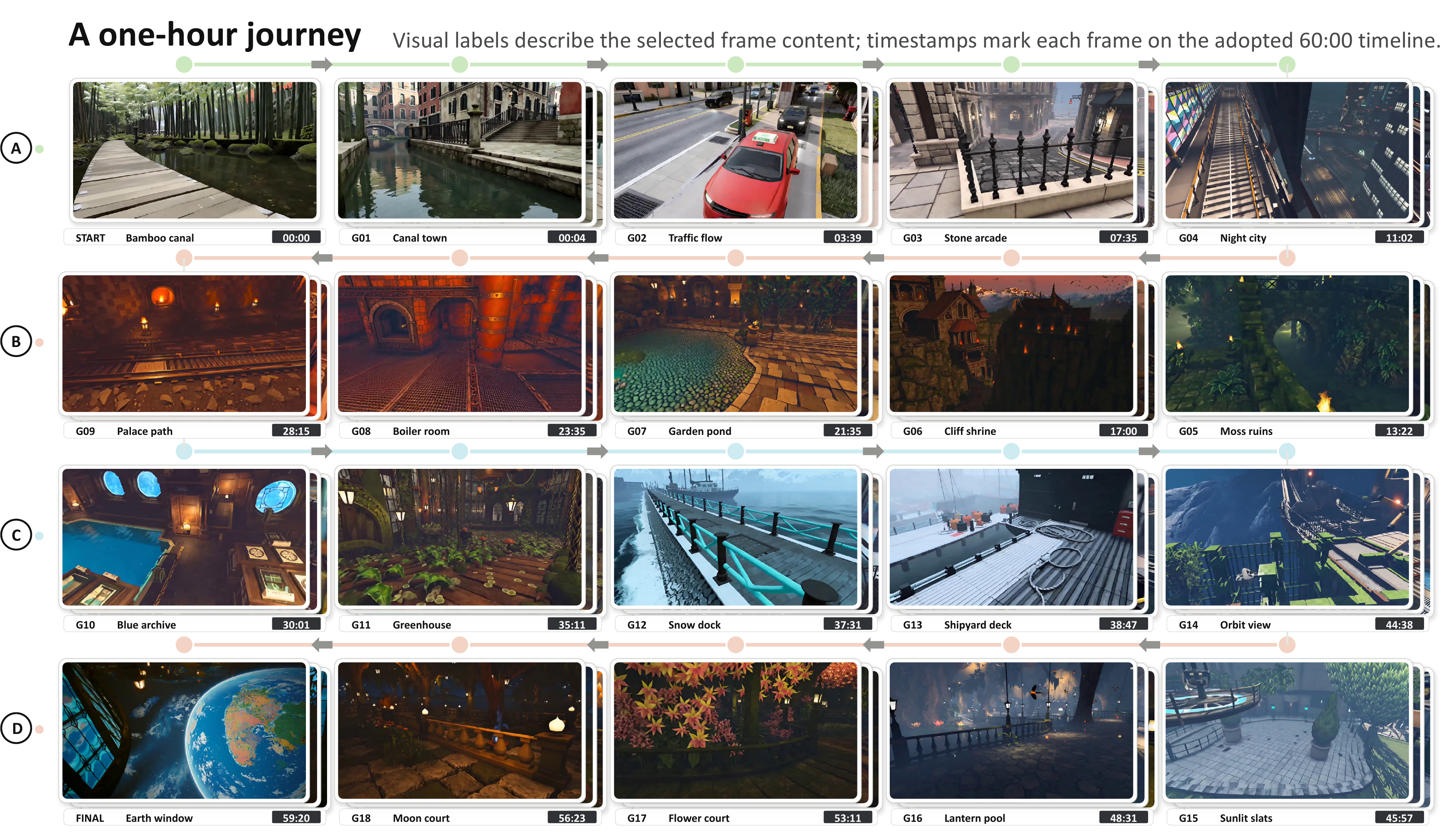}
\caption{
Hour-long world rollout. We sample frames from a single 60-minute generated session, covering 20 distinct scenarios. The timeline shows that the model can maintain coherent scene structure, visual quality over an extended uninterrupted rollout, providing a qualitative stress test for long-horizon stability.
}
\label{fig:long_demo}
\end{figure*}

\begin{figure*}[!ht]
\centering
\includegraphics[width=0.98\linewidth]{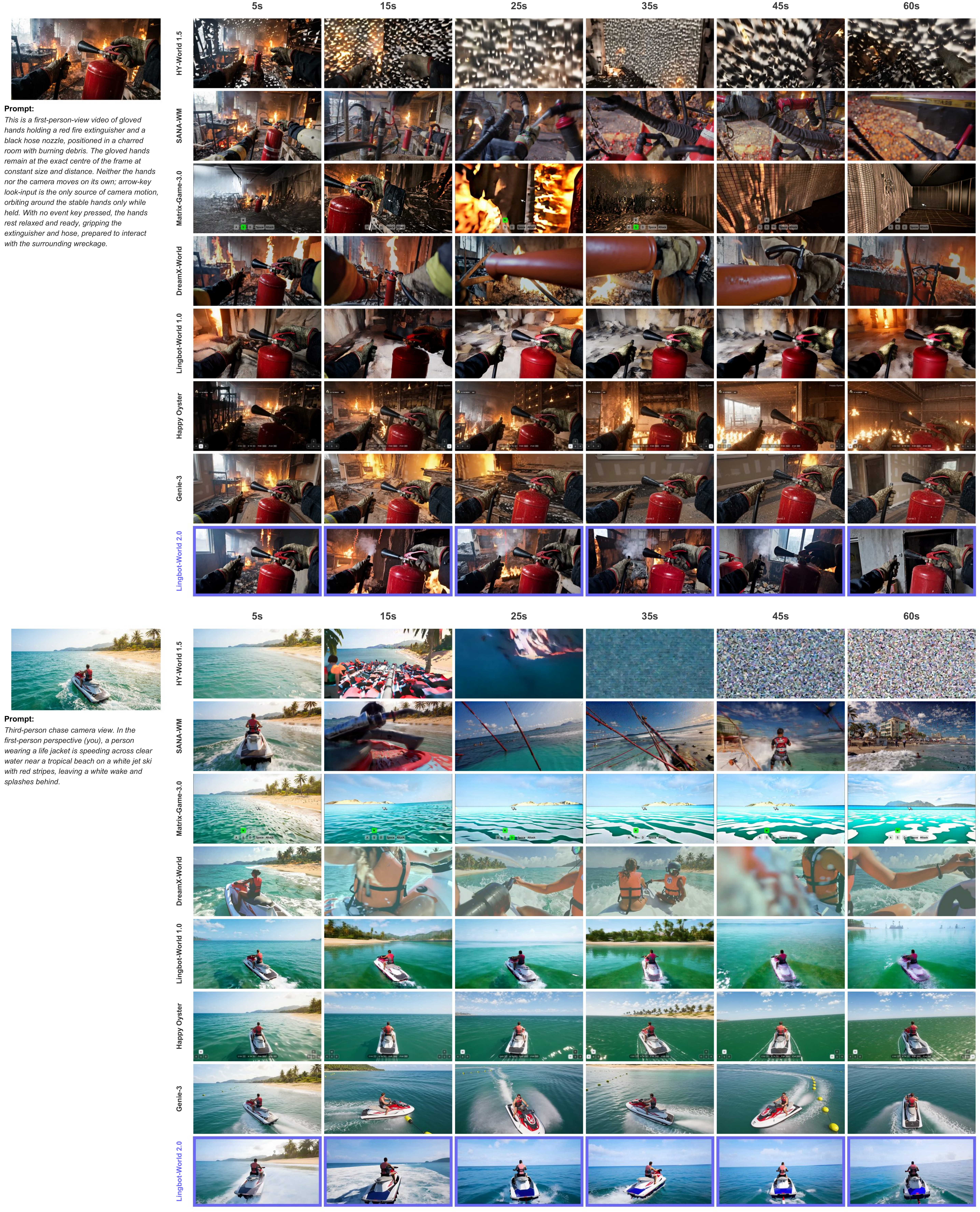}
\vspace{-6pt}
\caption{Qualitative comparisons. Our model maintains stable visual and physical consistency over long-horizon generation.}
\label{fig:qualitative1}
\end{figure*}

\begin{figure*}[!ht]
\centering
\includegraphics[width=0.98\linewidth]{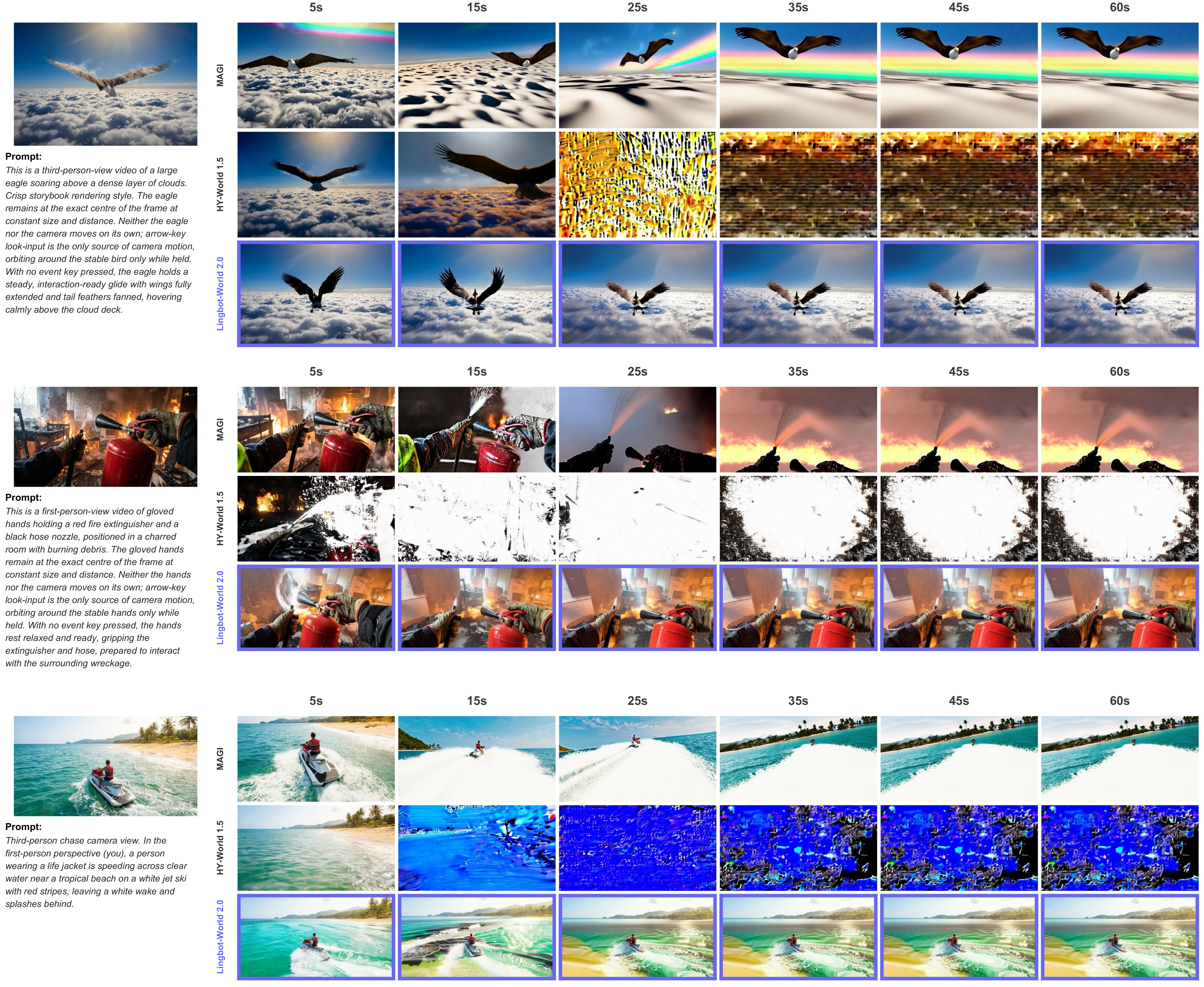}
\vspace{-6pt}
\caption{Qualitative comparisons on causal pretraining. Our model shows stable performance compared with baselines.}
\label{fig:pretrain}
\end{figure*}

\subsection{User Interface Design}

Figure~\ref{fig:ui} shows the interface of our interactive system. The layout is organized around three regions. The center hosts the live generation viewport, where the world is rendered in real time and responds immediately to user input.
The bottom panel exposes the low-level control scheme, using the WASD keys for movement and the IJKL keys for view control, so that camera navigation feels familiar to anyone accustomed to standard game controls. The right-hand panel is the agentic control surface, which lists the ``Event Proposals'' produced by the ``Director'' (VLM), each bound to a hotkey.

The event hotkeys fall into three groups. First, a set of \textbf{fixed keys} is always available regardless of context: \emph{Space} triggers a jump and \emph{P} triggers a wing glide. Second, the VLM continuously proposes \textbf{context-aware events} tailored to the current scene, split into
character actions bound to \emph{U} and \emph{O} (e.g., melee and ranged actions performed by the controllable subject) and environmental changes bound to \emph{F} and \emph{G} (e.g., a cloud whale passing by or a twilight descent).
Third, the remaining numeric keys \emph{(1, 2, 3, \dots)} are
\textbf{user-registered slots}, into which the user can enroll arbitrary events they wish to see, giving them direct authorship over how the world evolves. In the example shown, the controllable subject is an eagle gliding above a sea of
clouds, and the user has registered events such as a steep dive toward the clouds, an incoming storm, and a hard banking turn on keys 1 through 3.
\section{Results}\label{sec:results}

In this section, we evaluate our models along two axes. We first assess the real-time \textbf{causal distilled} model, which is the system intended for interactive deployment, comparing it against both proprietary and open-source baselines and examining the capabilities that emerge at long horizons. We then
report results for the \textbf{causal pretrained} backbone, which serves as the teacher for distillation.

\subsection{Causal Distilled Model}
\label{subsec:distill}

\paragraph{Comparison with prior work.}
We benchmark our distilled model against state-of-the-art closed-source systems HappyOyster \cite{happyoyster} and Genie 3 \cite{genie3}, as well as leading open-source interactive
world models \cite{zhu2026sana,matrix3,team2026dreamx,lingbot-world,Worldplay}. We evaluate along the dimensions most relevant to an interactive setting, namely visual fidelity, temporal stability over long rollouts, responsiveness to actions, and real-time throughput. As shown in Fig.~\ref{fig:qualitative1}, our model matches or exceeds the visual quality of the strongest closed-source baselines while remaining fully open, and it is the
only system in our comparison that sustains high-resolution generation in real time without visible degradation. Besides, our model supports much more interactions. 

\paragraph{Long-horizon stability.}
Beyond aggregate benchmark numbers, a key property of an interactive world model is its ability to remain stable over extended rollouts. To probe this limit, we generate a single uninterrupted session of over one hour, as shown in~\cref{fig:long_demo}. Throughout the run, visual quality shows no perceptible decay, and the world remains coherent and explorable from start to finish. This confirms that the model's stability is a structural property rather than a short-lived effect observed only on favorable clips.

\paragraph{Additional results.}
We further present qualitative results illustrating the breadth of the model's action space and its behavior across diverse scenes and interaction patterns in Fig. \ref{fig:result1} and Fig. \ref{fig:result2}.

\subsection{Causal Pretrained Model}
\label{subsec:pretrain}

We now evaluate the causal pretrained backbone that underpins the distilled model above. We compare it against prior causal world models under matched settings, focusing on visual fidelity and, in particular, long-horizon stability. As shown in Fig.~\ref{fig:pretrain}, our backbone consistently outperforms competing models \cite{MAGI-1,Worldplay}. While prior causal models degrade within seconds to a few minutes as errors accumulate, ours retains sharp textures, stable geometry, and a coherent scene over substantially longer rollouts. We attribute
this advantage to our anti-drift training, which explicitly discourages the model from compounding its own errors.
\section{Conclusion and Discussion}
\paragraph{Conclusion.}
We present \method, an open causal video generation
model for interactive world modeling. It pairs state-of-the-art visual quality with strong resistance to drift, and from it we distill a real-time model that sustains an unbounded, drift-free world at 720p and 60\,fps. Beyond navigation, the model supports a rich action space including combat, archery, spell-casting,
and shooting, along with on-the-fly environmental changes such as snow and rain. Finally, a pilot and director agentic harness orchestrates the model into a self-sustaining, goal-directed, and open-ended experience. We release our models and harness to the community.

\paragraph{Limitations and discussion.}
Several limitations remain. The most fundamental is \textbf{long-term memory}.
While the model stays visually stable over long horizons, it does not truly remember the world it has generated: a region that leaves the context window and is later revisited tends to be regenerated rather than recalled, so the world is persistent in appearance but not in identity. We are exploring mechanisms such as dynamic KV-cache scheduling to retain a larger and more relevant history, but this faces an inherent tension, since memory cost grows with the horizon while the available budget is bounded. Giving interactive world models genuine, compact long-term memory remains an open problem. A second limitation concerns \textbf{identity and style consistency}. Over very long rollouts, specific characters can subtly change in appearance and the overall art style may gradually drift, even when the scene as a whole stays coherent. A third limitation lies in \textbf{physical understanding}. The model learns dynamics purely from pixels, without an explicit notion of geometry or collision, so its grasp of the underlying physics is imperfect. As a result, physically implausible artifacts occasionally arise, such as characters or objects intersecting or passing through one another rather than colliding as they would in a real world. Endowing the model with a more faithful sense of physical interaction is an important direction for future work. Finally, although our distilled model runs in real time, it still requires substantial computational resources. Reaching real-time, high-fidelity world modeling on commodity hardware will need further gains in efficiency, which we see as an important direction going forward.
\section{Authors}\label{sec:authors}

\subsection{Core Contributors}

Zelin Gao$^1$, Qiuyu Wang$^1$, Jiapeng Zhu$^2$, Jingye Chen$^3$, Zichen Liu$^3$, Qingyan Bai$^3$, Jiahao Wang$^4$, Yufeng Yuan$^4$, Hanlin Wang$^4$, Yichong Lu$^4$, Yinghao Xu$^1$, Haojie Zhang$^2$, Tianrui Feng$^3$, Yuzheng Liu$^3$, Ka Leong Cheng$^4$, Jian Gao$^4$,  Yao Yao$^4$,  Xing Zhu$^5$, Yujun Shen$^5$, Hao Ouyang$^{\dagger}$

\vspace{5pt}
\noindent\textit{Superscripts denote contribution areas: \textsuperscript{1}Pre-training, \textsuperscript{2}Post-training, \textsuperscript{3}Deployment, \textsuperscript{4}Data, \textsuperscript{5}Project sponsors, \textsuperscript{$\dagger$}Project lead.}

\subsection{Contributors}
Yipengjing Sun, Liangxiao Hu, Yue Yu, Yihang Chen, Zikun Dai, Leyi Xu, Jiayi Zhu, Yihao Meng, Yanhong Zeng, Yangyan Li 
\section*{Acknowledgments}
We thank
Xiaoyue Duan, 
Yongtao Huang, 
Bo Jiang, 
Yibo Lu, 
Xiaoxi Ma, 
Shengnan Xu, 
Jingmei Zhao, 
Shuai Zhou, 
Yida Zou 
(\textbf{\textit{listed alphabetically by last name}}) for their valuable discussions and assistance.
{
\small
\bibliographystyle{plain}
\bibliography{ref.bib}
}

\end{document}